\pdfoutput=1
\documentclass[11pt]{article}

\usepackage{EACL2023}
\usepackage{amsmath}
\usepackage{times}
\usepackage{latexsym}
\usepackage{booktabs}
\usepackage{microtype}
\usepackage[pdftex]{graphicx}
\usepackage{subfigure}
\usepackage{color,soul} 
\usepackage[T1]{fontenc}
\usepackage[utf8]{inputenc}

\usepackage{microtype}

\usepackage{inconsolata}

\title{Distinguishing Fictional Voices: a Study of Authorship Verification Models for Quotation Attribution}

\begin{NoHyper}

\author{Gaspard Michel$^{\dagger\ast}$ \\ \texttt{gmichel@deezer.com} \And Elena V. Epure$^\dagger$  \\ \texttt{eepure@deezer.com}\And
  Romain Hennequin$^\dagger$ \\ \texttt{rhennequin@deezer.com} \AND Christophe Cerisara$^\ast$ \\ \texttt{christophe.cerisara@loria.fr} \AND
  $^\dagger$ \normalfont{Deezer Research, Paris, France} \\ $^\ast$ \normalfont{Loria, Nancy, France}
 }
\end{NoHyper}

\begin{document}
\maketitle
\begin{abstract}
% Through dialogues, characters in novels reveal aspects of their personality to the reader.
Recent approaches to automatically detect the speaker of an utterance of direct speech often disregard general information about characters in favor of local information found in the context, such as surrounding mentions of entities.
In this work, we explore stylistic representations of characters built by encoding their quotes with off-the-shelf pretrained Authorship Verification models in a large corpus of English novels (the Project
Dialogism Novel Corpus).
Results suggest that the combination of stylistic and topical information captured in some of these models accurately distinguish characters among each other, but does not necessarily improve over semantic-only models when attributing quotes.
However, these results vary across novels and more investigation of stylometric models particularly tailored for literary texts and the study of characters should be conducted.

\end{abstract}

\section{Introduction}

In prose fiction, entire universes come to life.
Different techniques are employed by authors to create engaging narratives and use a combination of narrator and character words to build the atmosphere and unveil the story.
Characters in the fictional world reveal aspects of their personalities through dialogues.
In Bakhtin's idea of \textit{polyphony} \cite{c7aeddce-79c1-3680-85ac-4759df6473c3}, characters participate in dialogues in their own voice, according to their own ideas about themselves and the fictional world. 
Automatically identifying parts of dialogues and attributing them to the character that utters them is central to many studies of large literary corpora \cite{Elson2010, Muzny2017a, Sims2020}

\begin{figure}
    \centering
    \includegraphics[width=\linewidth]{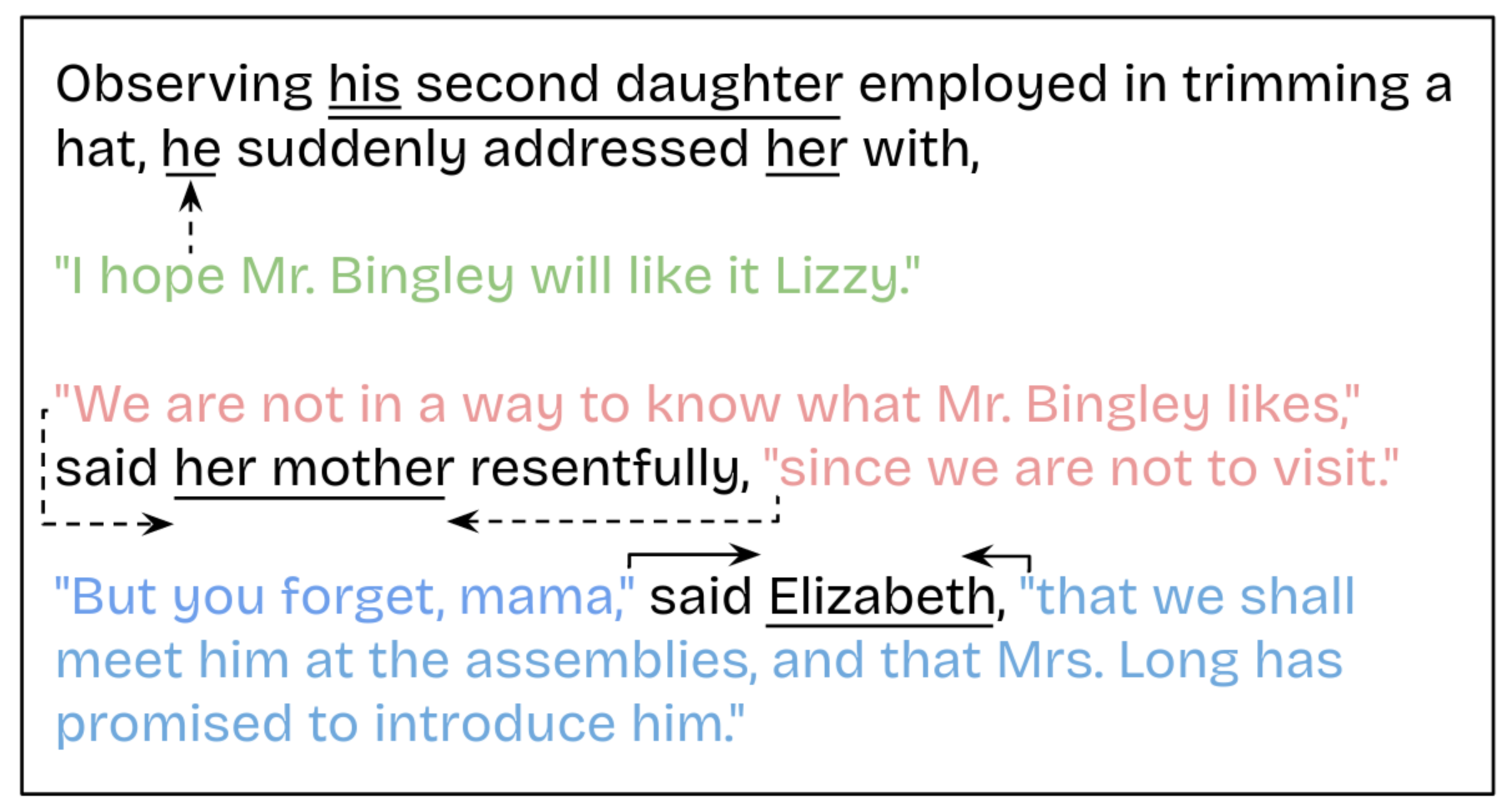}
    \caption{Example of quotation attribution on an excerpt of \textit{Pride and Prejudice} by Jane Austen (1813). Underlined text are identified mentions, and arrows link quotes to their relevant entity mention (solid arrows are explicit references and dashed arrows are anaphoric references). In a separate step, coreference resolution is used to link entity mentions to their canonical character.}
    \label{fig:quotation_attribution}
\end{figure}

The detection of direct-speech has been widely performed for English literature, and simple regular expression systems achieve almost perfect performances on well-formatted texts.
Attributing characters to quotes is more challenging and often requires solving multiple tasks: quotation identification, character identification and speaker attribution \cite{Muzny2017, Vishnubhotla2023}.
% Identifying characters and their mentions is generally done by extracting entity clusters with coreference resolution models.
A speaker is attributed to a quote by training a separate model to find the nearest relevant \textit{entity mention}, which is then linked to a \textit{canonical character} with coreference resolution models. 
Figure~\ref{fig:quotation_attribution} summarizes this process.
Although many approaches have been explored in this direction, there is still room for improvement

Using a recently proposed corpora of English novels annotated with speakers, our current work first investigates to which extent \textit{voices} of characters in novels are distinguishable using authorship verification approaches applied to character utterances.
Then, we analyze quotes of characters in this large corpus and evaluate to which extent character-related features encoded by pretrained authorship verification models contain a predictive signal for \textit{quotation attribution}.
Our intuition is that character-level information (such as style, preferences in topic, persona) might be used in addition to contextual information to improve quotation attribution models.
Prior stylometric studies have shown that canonical drama authors are able to create memorable characters with distinguishable voices \cite{Vishnubhotla2019, Sela2023}.
Nonetheless, the stylometric analysis of characters in novels remains scarce, mainly due to the lack of available corpora annotated with speakers.
To the best of our knowledge, exploring this type of character representations for quotation attribution has not been done before.

%Prior stylometric studies have shown that canonical drama authors are able to create memorable characters with distinguishable voices \cite{Vishnubhotla2019, Sela2023}.
%Nonetheless, the stylometric analysis of characters in novels remains scarce, mainly due to the lack of available corpora annotated with speakers.
%Using a recently proposed annotated corpora of English novels, our current work proposes to investigate to which extent \textit{voices} of characters in novels are distinguishable through the prism of authorship verification.
%Furthermore, we analyze quotes of characters in this large corpus and evaluate to which extent character-related features encoded by pretrained authorship verification models contain a predictive signal for quotation attribution.
%We hypothesize that such character-level information can be used in addition to contextual information to improve quotation attribution models.%and hope this study will encourage further research in this field. 

% Authorship verification of literary texts has mostly been used to understand stylistic differences of authors using for example grammatical features \cite{Baayen1996} or character n-grams \cite{Dinu2008}.
% However, the stylometric study of literary characters has received less attention, and has been mostly focused on characters in plays.
% Instead, we investigate characters voices in novels and design experiments to assess if a character representation built from a subset of its quotes can be used as a predictive signal for quotation attribution.

Consequently, with this work, we make the following contributions:
\begin{itemize}
    \item[1.] We investigate recent neural authorship verification models for the study of characters in novels and benchmark them on their ability to attribute authorship for distinguishing character voices in a large corpus.
    %\item[1.] We compare the capacity of two recently proposed authorship verification models to successfully distinguish voices of characters in 28 novels published between the 19th and 20th century.
    \item[2.] Framing quote attribution as an authorship verification task, we are the first work to evaluate the usefulness of stylometric character representations encoded by off-the-shelf authorship verification models to attribute quotes to characters.
\end{itemize}

Results suggest that most characters in the PDNC corpus own distinct voices, and that they are best distinguished by models that encode both semantic and stylistic information.
Semantic-only models, however, seem to be better at attributing quotes than models that encode style.
Besides, representing characters with the quotes they uttered in a single chapter appear to contain a predictive signal for attributing quotes in other chapters, but this varies per novel.
Finally, our results suggest that there are semantic variations between \textit{explicit} quotes (i.e. quotes where the relevant gold mention is a named mention of the speaker) and other type of quotes (\textit{anaphoric} and \textit{implicit} quotes, introduced in Section~\ref{sec:experiments}), and that including stylistic information alleviates the impact of these semantic shifts when distinguishing characters voices based on \textit{explicit} quotes only.
%Prethe l\inked textit{}viougold s computationmentionw here al studies seek to model \textit{dialogism} through the use of lexical features \cite{Burrows1987} or grammatical features \cite{Muzny2017Dialogism}.
\section{Related Work}
\label{sec:related}

% We further review works in quotation attribution, general and stylometric character representations as well as authorship verification.

\subsection{Quotation Attribution}
\label{sec:quotation_attr}

Quotation attribution models in novels often assume given utterances of direct speech.
\citet{Elson2010a} introduce the CQSC corpus and attribute automatically extracted quotes to named entities (``Elizabeth'') and nominals (``her daughter'') with a mention ranking model.
Instead, \citet{He2013} attribute quotes directly to \textit{speakers} with a supervised ranking system using features such as speaker alternation patterns and character-level features \cite{He2010}.
The deterministic sieve-based model of \citet{Muzny2017} regards quotation attribution as a two-step process: quote-\textit{mention} linking and mention-\textit{speaker} linking.

The NLP pipeline dedicated to books, BookNLP\footnote{\url{https://github.com/booknlp/booknlp} }, went a step further by replacing the deterministic sieves with fine-tuned language models.
\citet{Vishnubhotla2022} introduce the largest-to-date corpus of quotation attribution, PDNC, and show a similar accuracy score of around 63\% for both BookNLP and the sieve-based model.
However, better results were obtained later by fine-tuning BookNLP on PDNC \cite{Vishnubhotla2023}.
Although these works are considered state-of-the-art in quotation attribution, they inherently lack character-level information in the mention-speaker linking step.

\subsection{Character Representations}
\label{sec:related_char_repr}

%Representing characters through different aspects plays an important role for narrative understanding.
Most works focus on creating distributed \textit{embeddings} that encode the persona of characters (i.e.
characters with similar properties such as gender, job, relationships should have similar persona-based representations).
\citet{Bamman2014} propose a Bayesian model that infers latent character personas as a distribution over various dependency relations.
\citet{Brahman2021} introduce LiSCU, a dataset containing literary texts along with their summaries and descriptions of characters participating in the narrative.
They train a language model to generate accurate descriptions of characters, showing that the model has a complex understanding of personas.
\citet{Inoue2022} propose to represent characters in novels using a graph-based character network and positional embeddings.
The character network contains book-level and authorial information, and captures the attributes of characters, while positional embeddings encode the dynamics of character activity throughout the narrative.

In this work, we rather focus on \textit{what characters say} and \textit{how they say it}, building stylometric representations of characters with off-the-shelf pretrained authorship verification models.

\subsection{Authorship Verification}
\label{sec:av}

Authorship verification aims to predict whether two texts have been written by the same author. 
Recently, these models have employed a contrastive learning framework to build a representation space where works written by the same author are close together while being distant to texts written by other authors.
Evaluation is made by building disjoint sets of \textit{queries} and \textit{targets}.
Queries are pieces of text written by an author, and targets are other texts written by the same author and other authors.
Based on a similarity measure such as cosine similarity, a ranking distribution is created by scoring a query against all targets.
Area Under the Receiver Operating Characteristic Curve (AUC) is often used to evaluate if this distribution gives a high rank to the correct target.

Recent advances exploit language models to distinguish hundred of thousands of authors.
\citet{RiveraSoto2021} fine-tune SentenceBERT \cite{reimers-gurevych-2019-sentence} using thousands of Reddit users, Amazon reviews and Fanfiction stories.
Their model, LUAR, uses both stylistic and content information (such as topical preferences) to distinguish between authors.
Similarly, \citet{Wegmann2022} fine-tune RoBERTa \cite{Liu2019} on posts of thousands of Reddit users.
By controlling for content in the creation of their training data, they ensure that their model, STEL, mostly encodes stylistic information.

Although both models perform well on their respective authorship verification tasks, they are blackbox models that do not offer an interpretation of aspects of style captured in their representations.
\citet{Wegmann2022} present a clustering analysis of learned representations of Reddit posts and showed that STEL mostly captures variations of punctuation, casing and contraction spelling.
These stylistic variations do not apply to quotes in novel, we thus expect STEL to struggle to transfer from Reddit to our domain.
\citet{RiveraSoto2021} do not study aspects of style captured in LUAR's representations, but offer an interpretation of the model's performance based on the domain it was trained on.
Particularly, they show that LUAR is prone to overfit to the training domain style features.
While training on Reddit data, they conclude that the model rely less on topical diversity to distinguish among authors, which can be favorable to distinguish novel characters that usually speak in a wide range of topics.  

\subsection{Stylometric Analysis of Characters}

Stylometric analysis of literary characters has been mostly focused on drama characters because of existing large annotated corpus.
Most works focus on the style of character, aiming at capturing syntactic, lexical and phonological variations that can occur when they are quoted.
\citet{Vishnubhotla2019} propose to study the distinctiveness of character stylistic and topical patterns with text classification.
%They trained a supervised model to attribute quotes to characters using diverse set of features, and concluded that it is easier to distinguish character voices in the work of canonical authors.
\citet{Sela2023} propose a measure of distinctiveness based on character 3-grams, which they
apply to a large number of drama characters.
They show that it is able to capture interesting aspects of stylistics such as phonological differences, accents and dialects, as well as topical and lexical differences.
To the best of our knowledge, \citet{Dinu2017} is the only work that focuses on novel characters.
%Classification accuracy is also used as a measure of character distinctiveness in the epistolary novel ``Liaisons Dangereuse''.
Their supervised bag-of-word classification model was able to accurately classify some characters, but fell short on the main character of the epistolary novel ``Liaisons Dangereuse''.

Other related works leverage dialogues in movie scripts to build character representations.
\citet{Azab2019} train a Word2Vec model where the context window consists of the surrounding speaker identities as well as the current speaker utterance.
%By learning to predict the current speaker identity, the proposed model constructs fine-grained representations that encode both stylistic and topical information.
Similarly, \citet{Li2023} 
%use movie scripts and learn character representation from conversations.They 
encode all utterances of a script with a pre-trained language model, and extract representations by pooling all encoded quotes of a character together.
A contrastive learning objective is used to create a fine-grained representation space where characters are well separated.
%Their method improved on character-related tasks such as coreference resolution or character linking.
Although the methods presented above are quite similar to the way we build character representations, authors did not release the code publicly at the time of writing, precluding comparison in our experiments. 

In this work, we analyze novel characters at a larger scale using the PDNC corpus containing 28 English novels.
Instead of employing classification accuracy as a measure of character distinctiveness, we frame the task as an authorship verification problem to evaluate to what extent characters voice can be distinguished.
%Character representations are built with pretrained authorship verification models and used to quantify the distinctiveness of character voices.
We are also the first to evaluate if these character-level features contain a predictive signal to attribute unseen quotes to the right speaker.

%We use both LUAR and STEL to create these stylometric representations.
%\hl{Although authorship verification models have been shown to transfer poorly to other domains, we believe they still form a good baseline for evaluating the style of characters in novels.} \cite{rivera-soto-etal-2021-learning}

%In this work, rather than designing a new quotation attribution approach, we propose to study topical and stylistic representations of characters and evaluate to which extent they contain enough information to enhance modern attribution systems.

\section{Experimental Setup}
\label{sec:experiments}

% We focus on representations built from \textit{explicit} quotes, when the gold \textit{mention} linked to the quote is a named mention of the character.
% Explicit quotes are useful because they are easily attributed to the correct character \cite{Muzny2017, Vishnubhotla2022}, and can thus be used as a partial signal of a character's voice.

Our goal in this work is to investigate if fictional voices of literary characters in novels are distinguishable from a stylistic point-of-view.
We also want to know if a partial signal of a character's voice derived from its \textit{explicit} quotes (i.e. the gold \textit{mention} linked to the quote is any named mention such as ``Elizabeth'') is a good proxy for its overall voice.
Explicit quotes are straightforward to attribute to characters since they are linked to a named mention, which can then be linked to canonical characters (\textit{e.g.} with coreference resolution or name clustering) more easily than when dealing with pronominal mentions \cite{Muzny2017}.
We hypothesize that if we can construct representative character embeddings based on explicit quotes only, then these representations can in turn enhance quotation attribution solutions to detect the speaker of other type of quotes.
Other types of quotes include \textit{anaphoric} quotes (i.e the gold \textit{mention} is a pronoun or noun phrase) and \textit{implicit} quotes (often happens during a conversation, when no \textit{mention} is linked to the quote but the speaker can be inferred from the context).
Finally, using the same set of character representations, we want to evaluate to which extent they contain information to attribute quotes that were not used to build the representations. 

To evaluate these representations, we formulate the task as the authorship verification task: given a corpus of quotes from character A (the \textit{query}), a corpus of other quotes from character A and similar corpora for other characters in a given novel (the \textit{targets}) and a similarity measure, we evaluate the ability of pretrained models to predict if the targets have been written by character A or not.
AUC is used to assess models' performances, as it accounts for how well models can rank predictions, without concerns of threshold values \cite{Tyo2022}.
We chose to frame the task as an authorship verification problem rather than closed-set authorship attribution because the number of targets (\textit{i.e.} number of candidate speakers) vary for each query, which is further described in Section~\ref{sec:eval}
%are pretrained models able to give a higher rank to the target of character A than to others targets ?

We first present how character representations are derived from pretrained models, and then describe how we evaluate the capacity of these representations to answer the above questions. We publicly release our code for further research\footnote{\url{https://github.com/deezer/quote_AV}}.

\subsection{Building Character Representations}
\label{sec:char_repr}

Transformer-based models are widely used to encode textual information.
To build character representations, we leverage various publicly available pretrained models (PM) trained on different tasks as quote encoders.
For each novel, we assume that we have access to all utterances of direct speech $Q = \{ q_1, \dots, q_n \}$ as well as each character in the novel $C = \{ c_1, \dots, c_m\}$.
Let $g : Q \mapsto C$ be a function that assigns a quote $q_i$ to its speaker $c$ such that $g(q_i) = c$ implies that character $c$ is the speaker of the quote $q_i$.
% The speaker of a quote, $s_i$, is supposed to be one of the characters in the novel.
% Therefore, $s_i = c_j$ implies that character $c_j$ is the speaker of the quote $i$. 
To build the representation of a character $c$ in a given subset of quotes $\tilde{Q} \subset Q$, we first extract all quotes of character $c$ in the subset: $\tilde{Q}_c = \{ q_i : q_i \in \tilde{Q}, g(q_i)=c\}$.
A quote representation is obtained by encoding each quote with a pretrained model, denoted as $\text{PM}_\theta$: 
$$
\textbf{h}_{q_i} = \text{PM}_\theta ( q_i ) 
$$
We then derive an embedding of character $c$ in the subset $\tilde{Q}$ by pooling all embeddings of quotes spoken by $c$ in $\tilde{Q}$:
$$
\textbf{H}_{\tilde{Q}_c} = \text{POOL}(\{\mathbf{h}_{q_i}: q_i \in \tilde{Q}_c\})
$$
In our experiments, the $\text{POOL}$ function is the average of all quote representations, except for the LUAR model that uses an attention-based $\text{POOL}$ function with attention weights trained to focus on the relevant texts of an author.  
By pooling over the subset of quotes of a character, we expect the resulting representation to contain general information of \textit{what a character say} and/or \textit{how he says it}, depending on the PM used.
Compared to some of the previous approaches to character representations presented in section \ref{sec:related_char_repr}, we do not use any contextual information (surrounding passages of narrative text, sequence of speaker turn, or surrounding quotes) so that that the representations focus mainly on stylistic and/or content information. We conduct different experiments by varying the construction of the subset $\tilde{Q}$.

\noindent\textbf{Chapterwise}:
we extract all quotes of a character in a given chapter $T$ to build its query representation. The targets are created by using quotes contained in the held-out chapters.   

\noindent\textbf{Explicit}:
we only extract \textit{explicit} quotes of a character in a given chapter $T$ with similar targets as in the chapterwise experiment.
We thus build representations for a character with quotes that are linked to a named mention of the character.
This experiment is designed such that we can quantify the amount of information lost compared to the chapterwise experiment that uses all types of quotes. 
It might happen that some characters are not explicitly quoted in chapter $T$.
In this case, we do not build representations for these characters.

\noindent\textbf{Reading Order}:
we use the first $n$ quotes of a character in the first half of novels (segmented by chapter) as a basis of its query representation.
Targets are built using quotes in the remaining half.
With this experiment, we want to see the impact of increasing the amount of available character information on the capacity of models to distinguish their voices.

\subsection{Data}
\label{sec:data}

We use the PDNC dataset\footnote{\url{https://github.com/Priya22/project-dialogism-novel-corpus}} \cite{Vishnubhotla2022}, containing annotations of speakers at the quote level for 28 English novels written by 21 authors and published between the 19th and early 20th century.
This dataset consists of mostly literary novels, and a few children, crime and science-fiction novels.
Characters in each novel are labelled with \textit{minor}, \textit{intermediate} and \textit{major} roles, depending on the total number of quotes they uttered.
We only focus on \textit{intermediate} and \textit{major} characters that uttered at least 10 and 100 quotes respectively, and discarded \textit{minor} characters that participate less in the narrative.
Quotes are often subject to \textit{incises}, where a narrative segment giving indication on who and how the quotes is being said is inserted within the quote (\textit{e.g.} ``said her mother resentfully'', third paragraph in Figure~\ref{fig:quotation_attribution}).
In this case, we use the full text of the quote, discarding the incise, as a single character's utterance.

We build character embeddings using the methodology explained in Section~\ref{sec:char_repr} that we further use to derive sets of queries and targets.
For a character $c$ and its quote subset $\tilde{Q}_c$, the associated query is the character representation built from the subset, $\mathbf{H}_{\tilde{Q}_c}$.
Using the held-out subset, $O$, the associated set of targets are embeddings of every character that utters quotes in $O$: \{$ \mathbf{H}_{O_{c'}}$: \; $c' \in C \}$.
We only construct queries for characters that utter at least 5 quotes in $\tilde{Q}_c$ to mitigate the amount of uninformative queries.
We chose to use 5 quotes based on preliminary results showing that some queries would have only 1 quote and that the resulting character representations were not really informative.
Results of the reading order experiment presented in Section~\ref{sec:reading_order} further support this observation. 

Table~\ref{tab:summary_stat} summarizes the main statistics of the resulting data.
For the chapterwise experiment, we derived 1606 queries on the entire corpus, but only 562 for the explicit experiment.
Indeed, explicit quotes represent around 31\% of the total number of quotes in the corpus, with large discrepancy across novels (the minimum is 10\% and the maximum is 81\%), thus leading to many characters that do not utter at least 5 explicit quotes in $\tilde{Q}_c$.
As a result, the percentage of active characters (i.e characters that have at least one query) drops from 93\% to 53\% and, out of the 28 novels, we could not create queries for two novels because they did not contain enough explicit quotes to build representations (\textit{The Gambler} by Fyodor Dostoevsky, 1887, and \textit{The Sport of the Gods} by Paul Laurence Dunbar, 1902).
Queries in the Chapterwise experiment also contain twice as many quotes on average than in the Explicit one.  
Nonetheless, the number of character targets and the number of quotes in target is roughly the same between the two experiments.
We thus expect the task of distinguishing voices and attributing unseen quotes based on representation of explicit quotes only to be generally harder.  

\begin{table}
    \centering
    \begin{tabular}{l|c|c}
    \toprule
    & \textbf{Chapterwise} & \textbf{Explicit}\\
    \midrule
     Total queries & 1606 & 562 \\ 
    \midrule
     \# Speakers & 11.1 (\small{4.6}) & 11.1 \small{(4.6)}\\
     Activity (\%)  &  93 \small{(10)} & 53 \small{(28)}\\
     Queries & 57.4 \small{(29.3)} & 21.6 \small{(17.9)}\\
     Query length & 21.4  \small{(11.5)} & 10.5 \small{(3.5)}\\
     Targets/query & & \\
     \hspace{0.5cm} Character & 11.0 \small{(4.6)} & 11.2 \small{(4.7)} \\
     \hspace{0.5cm} Quote & 1142 \small{(600)} & 1176 \small{(597)} \\
     \bottomrule
    \end{tabular}
    \caption{Summary statistics of our set of queries and targets on the PDNC corpus. Bottom part is averaged over novels with (standard deviation).}
    \label{tab:summary_stat}
\end{table}

\subsection{Models}

We build representations with two pretrained authorship verification models: STEL and LUAR and introduce two baseline models: SentenceBERT (SBERT)\footnote{We use \texttt{all-mpnet-base-v2}} \cite{reimers-gurevych-2019-sentence} and a RoBERTa-based multi-label emotion classification model\footnote{\url{https://huggingface.co/SamLowe/roberta-base-go_emotions}}. 

\subsubsection{Baselines} 
The SBERT and Emotion models are referred to as baselines because their purpose is not to encode stylistic information of characters.
SBERT is trained to recognize semantically similar sentences, hence encoding rich semantic textual information.
We expect this semantic-only model to distinguish characters voices based on the content of their corpus of quotes, such as topical preferences.
The Emotion model is a RoBERTa model fined-tuned to classify emotions in a multi-label setup (28 emotions), allowing predictions of multiple emotions conveyed at once in the same sentence.
We use this model as a benchmark based on prior analysis we made, showing that some characters were conveying certain emotions more than others. Thus, our intuition was that it could be used as a discriminative feature of a character's voice.
We use representations contained in the last RoBERTa transformer layer before the classification head when encoding quotes. 

\subsubsection{Authorship Verification Models}

Authorship verification models are trained to predict if two texts (or corpus of texts) have been written by the same author, enabling them to capture authorial style to some extent.
STEL is a RoBERTa model fine-tuned on the Contrastive Authorship Verification (CAV) task with millions of Reddit users.
In the CAV task, a model is asked to distinguish which from three pieces of texts (triplets) have been written by the same author.  
Using triplets that have similar topic, STEL is trained to distinguish between authors using stylistic information only.
Although its base model, RoBERTa, encodes semantic to some extent, restraining training triplets to texts that essentially cover similar topic forces the model to focus on stylistic cues.

LUAR fine-tunes SentenceBERT to encode corpora of utterances.
Unlike STEL, they do not force training examples to have similar topic, leading to a model that encodes both content and stylistic information in author representations.
For the same author, different representations are built using distinct collections of documents written by the author.
LUAR encodes stylistic information by being trained on the authorship verification task.
Compared to STEL, we expect LUAR to build more robust character representations since it uses an attention mechanism that allows focus on texts with strong authorial signal.

For all models, we use a maximum sequence length of 64, truncating longer quotes (only 14\% of the total number of quotes are longer than 64 tokens).
We use publicly available versions of LUAR\footnote{\url{https://huggingface.co/rrivera1849/LUAR-MUD}} and STEL\footnote{\url{https://huggingface.co/AnnaWegmann/Style-Embedding}}. 

% The first three models are implemented in the \texttt{Sentence-Transformer}\footnote{\url{https://www.sbert.net/index.html}} package, and use the Huggingface implementation of LUAR\footnote{\url{https://huggingface.co/rrivera1849/LUAR-MUD}}.
% SentenceBERT (SBERT) encodes semantic information of quotes and is used as a baseline.
% We use the emotion classification model to assess if character voices can be distinguish by the emotion they convey.
% We check if stylistic features of characters are useful to distinguish their voices by testing both STEL and LUAR representations.
% For all models, we use a maximum sequence length of 64.

\subsection{Evaluation}
\label{sec:eval}

\subsubsection{Character-Character}
Given a query character representation built from a subset of quotes $\mathbf{H}_{\tilde{Q}_c}$, a similarity function $\phi$, and the held-out subset $O$, we evaluate the similarity of the query with the targets built from $O$.
In this work, we use cosine similarity for the $\phi$ function.
Ideally, we want a high similarity between a character query and the target linked to the same character, $\mathbf{H}_{O_c}$, and low similarity between the character query and other characters target $\mathbf{H}_{O_{c'}}$: 
\begin{equation}
\label{eq:char_char}
\phi(\mathbf{H}_{\tilde{Q}_c} , \mathbf{H}_{O_c} ) > \phi(\mathbf{H}_{\tilde{Q}_c} , \mathbf{H}_{O_{c'}} ) , \; \forall \; c' \neq c%, \; c' \in C     
\end{equation}
In practice, we evaluate the capacity of pretrained models to give a high rank to corresponding character target $\mathbf{H}_{O_c}$ using AUC.
In this context, the AUC measures the probability that Equation~\ref{eq:char_char} holds when randomly selecting a character $c'$ different than $c$.
We chose AUC over standard authorship attribution metrics (such as macro-averaged accuracy) because of its ability to evaluate the output ranking distribution, $\{ \phi(\mathbf{H}_{\tilde{Q}_c} , \mathbf{H}_{O_c'} ),  c' \in C$\}.
Besides, unlike accuracy, AUC does not require a threshold value for predicting the speaker of a quote, which can be tricky when using cosine similarities.
We refer to this evaluation as Character-Character (CC) as it measures how unique are characters voices. 

\subsubsection{Character-Quotes}

We now introduce how we evaluate the performances of such character representations at attributing quotes from the held-out subset.
Similar query representations are used, but targets are replaced by quote representations (encoded by the same PM) rather than character representations.
Let $q_i \in O_c$ be a target quote from character $c$ in the held-out subset $O$ and $q_j \in  \bar{O}_c = \bigcup_{c' \neq c} O_{c'}$ be any quote spoken by a different character in $O$, we evaluate the following hypothesis:
\begin{equation}
\label{eq:char_quote}
\phi(\mathbf{H}_{\tilde{Q}_c} , \mathbf{h}_{q_i} ) > \phi(\mathbf{H}_{\tilde{Q}_c} , \mathbf{h}_{q_j} ), \; \forall \, q_i \in O_c, \, q_j \in \bar{O}_c
\end{equation}
We also use AUC in this Character-Quote evaluation setup (CQ) to assess how well the target quotes spoken by character $c$ are ranked compared to quotes of other characters.
Here, the AUC measures the probability that Equation~\ref{eq:char_quote} holds when randomly selecting a quote $q_i \in O_c$ and a quote $q_j \in \bar{O}_c$.
Intuitively, a high AUC indicates that character representations are more similar to quote representations of the same character, thus showing that they contain useful information to attribute the right speaker to quotes.

%%%%%%%%%%%%%%%%%%%%%%%%%%%%%%%%
\begin{table}[t!]
\centering
\begin{tabular}{l|cc}
\toprule
 & \textbf{CC} & \textbf{CQ} \\
\midrule
Semantics & 67.3 \small{(11.6)} & \textbf{55.1} \small{(2.5)}  \\
STEL & 58.1 \small{(8.3)} & 52.8 \small{(1.9)}  \\
Emotions &  56.0 \small{(8.0)} & 51.7 \small{(1.5)}  \\
LUAR & \textbf{81.6} \small{(6.2)} & 53.6 \small{(2.4)}  \\
\bottomrule
\end{tabular}
\caption{AUC results of the \textbf{chapterwise} experiment. Results are averaged over novels (standard deviation). Best results are highlighted in \textbf{bold}.}
\label{tab:chapterwise_res}
\end{table}

\begin{table}[t!]
    \centering
\begin{tabular}{l|cc}
\toprule
 & \textbf{CC} & \textbf{CQ} \\
\midrule
Semantics & 63.9 \small{(15.8)} & 54.4 \small{(4.6)}  \\
STEL & 56.2 \small{(15.6)} & 52.7 \small{(3.6)}  \\
Emotions & 53.4 \small{(14.4)} & 51.4 \small{(3.1)}  \\
LUAR & \textbf{80.1} \small{(10.0)} & 53.5 \small{(4.4)} \\
\bottomrule
\end{tabular}
\caption{AUC results of the \textbf{explicit} experiment. Results are averaged over novels (standard deviation). Best results are highlighted in \textbf{bold}. Results for the CQ evaluation are not highlighted because the large standard deviations prevent to chose a best model.}
\label{tab:explicit_res}
\end{table}

%%%%%%%%%%%%%%%%%%%%%%%%%%%%%%%%

\section{Results}
\label{sec:results}

\subsection{Chapterwise}
Results for the chapterwise experiment are displayed in Table~\ref{tab:chapterwise_res}.
In the CC evaluation setup, semantic-only representations built from the SBERT model appear to be quite good at distinguishing the voices of characters.
We believe that SBERT particularly captures topical preferences, which appear as a useful discriminative feature of voices.
Nonetheless, purely stylistic information seems to be worse at distinguishing voices than semantic-only embeddings, as suggested by STEL results.
LUAR's high performance suggests that a combination of both content and stylistic information is desirable to achieve better and more stable discrimination among characters.
Overall, the Emotions model seems quite misleading, as the AUC is the closest to random attribution (a random attribution would lead to an AUC of 50\%).

When evaluating the capacity of these representations to attribute quotes, we see a drastically different picture.
The performance of all models is just slightly higher than random attribution, indicating that the task is generally harder.
This is not surprising, deciding which among thousands of quotes have been spoken by character $c$ given a corpus of around 10 quotes spoken by $c$ without access to contextual information is a challenging task, probably even for humans.
Interestingly, the semantic-only baseline achieve the best results here.
We hypothesise that the drop of performance of LUAR is mostly due to how it encodes quotes: it was trained to produce fine-grained author representations based on a corpus of multiple texts rather than to build rich text representations.
In contrast, SBERT directly produces meaningful quote embeddings, leading to better performance for quotation attribution even if resulting character representations are less informative than LUAR's.

The high standard deviation in these results also suggests that distinguishing voices of characters is easier in some novels than in others.
We analyze to which extent the semantic model and LUAR complement each other by looking at performance per novel in Appendix~\ref{sec:appendix_A} and per character role in Appendix~\ref{sec:appendix_B}.

%%%%%%%%%%%%%%%%%%%%%%%%%%%%%%%%%

\begin{figure*}[t]
    \centering
    \subfigure{\includegraphics[width=0.49\linewidth]{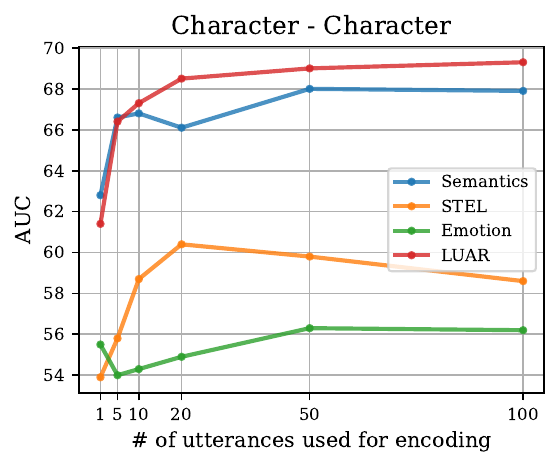}
    }
    \hfill
    \subfigure{\includegraphics[width=0.49\linewidth]{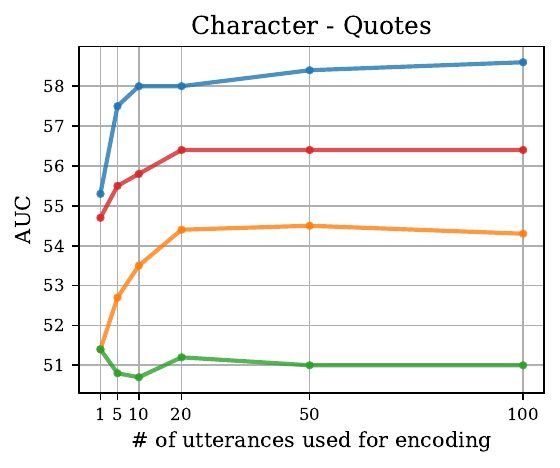}
    }
\caption{Results of the reading order experiment for the CC (left) and the CQ (right) evaluations. We look at AUC performance when varying the number of utterances used to create character representations.}
\label{fig:reading_order}
\end{figure*}

%%%%%%%%%%%%%%%%%%%%%%%%%%%%%%%%%
\subsection{Explicit}
We present the results of the explicit experiment in Table~\ref{tab:explicit_res}.
As expected, the performance of all models is worse than in the chapterwise experiment.
Indeed, character representations built from explicit quotes have access to fewer quotes, reducing the amount of available information for each character.
In the CC evaluation setup, LUAR still performs best at distinguishing voices of characters, followed by the SBERT model.
Interestingly, even though queries are built with twice as few quotes on average than in the chapterwise experiment, we observe only a slight performance drop for LUAR and STEL.
This observation suggests that explicit quotes constitute a strong signal of characters voice.
However, we observe a larger drop for the SBERT model, indicating that there might be semantic variations between explicit quotes and other types of quotes.
We hypothesize that such variations should occur less in stylistic cues of quotes, which is further supported by the lower AUC drop of STEL and LUAR.
 
Evaluating with the CQ setup, we can draw similar conclusions.
Performances are worse than in the chapterwise experiment, and we see a larger AUC drop for the SBERT model than for models that encode style.
However, the semantic model seems to remain the best at attributing unseen quotes on average, but it's not true for all novels. 

Compared to the chapterwise experiment, we see very large standard deviations across novels.
This is not surprising, some novels contain only a very small number of explicit quotes, leading to a smaller amount of queries as explained in Section~\ref{sec:data}.
Although aggregated results suggest that the semantic model and the LUAR model still contain information for attributing quotes, when looking at results per novel, we observe that they can sometimes provide misleading attributions with AUC worse than 50\%, but also provide a good ranking of quotes in other novels (the highest AUC is of 68\% for \textit{The Age Of Innocence} by Edit Wharton). 
Interestingly, we also observe a larger performance gap between the two models on some novels, indicating that they are more complementary when using explicit quotes only. 

\subsection{Reading Order}
\label{sec:reading_order}

Results for the reading order experiment are displayed in Figure~\ref{fig:reading_order}.
Looking at the CC evaluation, we see that all models except Emotions have better performances when increasing the number of available utterances from 1 to 20.
% We believe that's because the emotion model is being mislead by more complex emotion profiles of characters built using more quotes.
Increasing the number of quotes always improves the LUAR model, which successfully creates more fine-grained representations when accessing additional quotes.
However, the STEL model peaks at 20 utterances, indicating that it can't really capture the style of characters with more quotes.
Interestingly, the semantic model performance only varies slightly when using 5, 10 or 20 utterances, suggesting we can build meaningful semantic representations with a small number of quotes.
This result further supports our hypothesis that the drop of performance between the Chapterwise and Explicit experiments is closely linked to semantic variations between explicit quotes and other types of quotes rather than simply due to lower query sizes.
Overall, LUAR and Semantics build more informative representations using an increasing number of quotes from, 20 to 50 quotes of a character. 

Results for the CQ evaluation show a similar trend, where the AUC of all models plateaus starting from 10 or 20 utterances, except for the Semantics that have increased performance with more data starting from 10 quotes.
We hypothesize that stylistic information and topical preferences of characters can thus be captured by these models with a fairly low amount of quotes.
A more complex understanding of characters does not always help to attribute quotes when using quote embeddings built with the same models, highlighting the need for additional contextual information. 

%%%%%%%%%%%%%%%%%%%%%%%%%%%%%%%%

\section{Discussion}
\label{sec:discussion}

We conducted experiments to understand how character representations built from explicit quotes could help to improve quotation attribution.
These quotes are particularly easy to attribute to characters and can thus be detected automatically to build \textit{informative} character representations that can serve as additional inputs to quotation attribution systems.  
Results presented above suggest that explicit quotes might be a good proxy for the voice of fictional characters and that semantic and stylistic information of quotes can help attribute quotes.
Nonetheless, we think that there might be semantic variations between explicit and other types of quotes and that adding stylistic information in representations of characters alleviates this shift.

Experiments conducted in this work are focused on \textit{intermediate} and \textit{major} characters, i.e. characters that participate more and shape the narrative.
Although they represent a large number of different characters, \textit{minor} characters often have less impact on the story and do not contribute significantly to the overall number of quotes that we want to attribute.
However, even with characters that are more quoted, we observed discrepancies in authorial patterns of explicit quoting.
While some authors quote all their characters explicitly at least 5 times in a chapter, some do not.
As a result, we could build queries for only 53\% of \textit{intermediate} and \textit{major} characters in the PDNC corpus.
When looking at whole novels rather than at chapters, only 11\% are explicitly quoted less than 5 times, among which 17\% are major characters.
Therefore, we can still build representations for a majority of characters, which motivated our work. 

We studied stylistic information encoded in two off-the-shelf pretrained authorship verification models, LUAR and STEL.
These models have been trained to distinguish thousands of authors of Reddit posts, and have been shown to transfer poorly to other domains \cite{RiveraSoto2021}. 
% When analyzing learnt representations of style in STEL, \citet{Wegmann2022} found that it was very sensitive to punctuation and casing variations (e.g. finishing a sentence without punctuation markers, using \textit{i} instead of \textit{I}).
Most novels do not contain stylistic traits captured by STEL, which probably explains why it is performing badly.
We were aware of this limitation at first, but decided to test the model as an off-the-shelf solution to obtain stylometric representations.
In the future, we plan to re-train a STEL-like model on literary texts such as drama.
LUAR encodes both semantic and stylistic information, it is thus hard to infer the dimensions of content and style it captures as well as their respective contribution to the task.
Its good performance on the Character-Character evaluation setup suggests that it gets dimensions of style that make sense in literature. 
More generally, interpretable authorship verification models \cite{Patel2023} are an interesting direction as they combine the performance of neural approaches with the interpretability of frequency-based methods.

The high standard deviations across novels indicate that the task of distinguishing voices of characters is easier in some novels than in others.
In the Chapterwise experiment (CC evaluation), the AUC of LUAR goes as low as 68\% and as high as 91\%.
Ideally, we would like to understand the reasons behind these variations: Are some authors better at creating memorable voices? Is it easier in a particular genre?
Interpretability and literary knowledge are key to answer these questions, that we leave for future work.

\section{Conclusion}
\label{sec:conclusion}

We presented a study of recent neural approaches to authorship verification applied to literary characters.
We designed three experiments to assess if such models can be used to create meaningful character representations and to assess if explicit quotes were a good proxy of a character's voice.
Our first evaluation focuses on the ability of these representations to distinguish characters, while our second quantifies the amount of information they contain to attribute unseen quotes.
Results at the character level suggest that their voices are better distinguished when using a combination of stylistic and semantic information.
Using style also helps to reduce the impact of the semantic shift observed between explicit quotes and other types of quotes.
When attributing quotes, our results suggest that adding stylistic information does not necessarily improve over semantic-only models.
We believe that the main cause is a poor domain transfer from Reddit to English novels.
In the future, we plan to further analyze representations built from models trained on movie scripts \cite{Azab2019, Li2023}, which we argue should contain stylistic patterns more similar to the ones found in literary works.
We also want to investigate how such representations can be incorporated into quotation attribution systems.
Finally, we believe our approach could be used at a larger scale to investigate which authors/genre are better at constructing unique voices for their characters.

\bibliography{anthology,custom, bilbio}
\bibliographystyle{acl_natbib}
\begin{table*}[ht!]
\small
\centering
\begin{tabular}{l|cccc|cccc}
\toprule
& \multicolumn{4}{c}{\textbf{Chapterwise}} & \multicolumn{4}{c}{\textbf{Explicit}} \\
\midrule
 & \textbf{CC (M)} & \textbf{CC (I)} & \textbf{CQ (M)} & \textbf{CQ (I)} & \textbf{CC (M)} & \textbf{CC (I)} & \textbf{CQ (M)} & \textbf{CQ (I)} \\
\midrule
Semantics & 62.9 \small{(15.6)} & \textbf{75.6} \small{(12.8)} & \textbf{53.1} \small{(3.6)} & \textbf{59.6} \small{(5.0)} & 58.0 \small{(18.3)} & \textbf{79.1} \small{(16.9)} & 51.8 \small{(3.8)} & \textbf{61.2} \small{(7.3)} \\
STEL & 55.4 \small{(14.6)} & 62.2 \small{(11.1)} & 52.2 \small{(3.1)} & 53.6 \small{(3.3)} &   52.5 \small{(18.9)} & 64.5 \small{(23.7)} & 51.5 \small{(3.7)} & 55.0 \small{(10.5)}  \\
Emotions &  53.1 \small{(15.1)} & 59.5 \small{(10.0)} &  50.2 \small{(3.2)} & 53.6 \small{(7.6)} & 49.9 \small{(18.2)} & 59.6 \small{(23.9)} & 50.2 \small{(3.2)} & 53.6 \small{(7.6)} \\
LUAR & \textbf{91.2} \small{(4.3)} & 63.0 \small{(12.7)} & 52.1 \small{(3.9)} & 56.6 \small{(4.9)} & \textbf{87.6} \small{(9.0)} & 57.6 \small{(25.1)} & 51.6 \small{(4.5)} & 58.5 \small{(7.3)} \\
\bottomrule
\end{tabular}
\caption{AUC results by character role for the Chapterwise and Explicit experiments. (M) means major and (I) intermediate.}
\label{tab:per_role}
\end{table*}

\appendix

\section{Performance per Novel}
\label{sec:appendix_A}
\begin{figure*}[b!]
    \centering
    \subfigure{\includegraphics[width=0.49\linewidth]{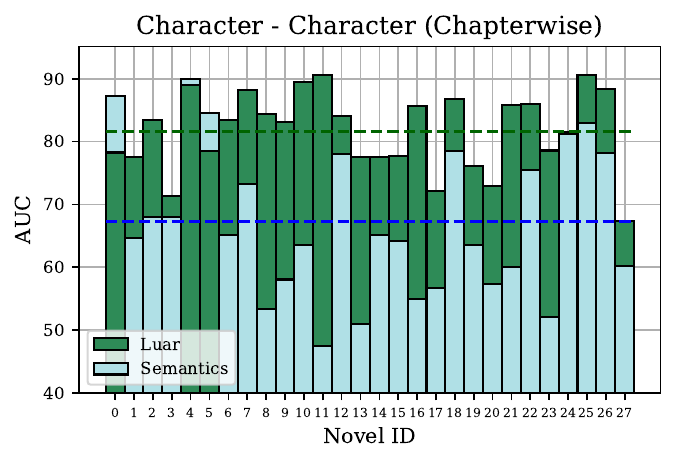}
    }
    \hfill
    \subfigure{\includegraphics[width=0.49\linewidth]{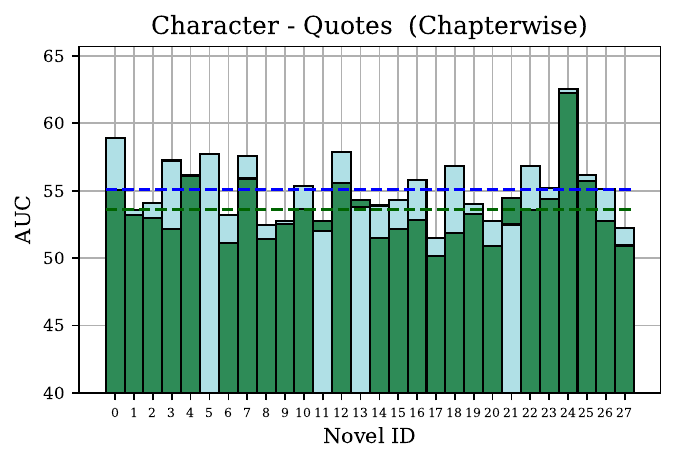}
    }
\caption{AUC per novel for the \textit{Chapterwise} experiment.}
\label{fig:chapterwise_novel}
\end{figure*}

\begin{figure*}[h!]
    \centering
    \subfigure{\includegraphics[width=0.49\linewidth]{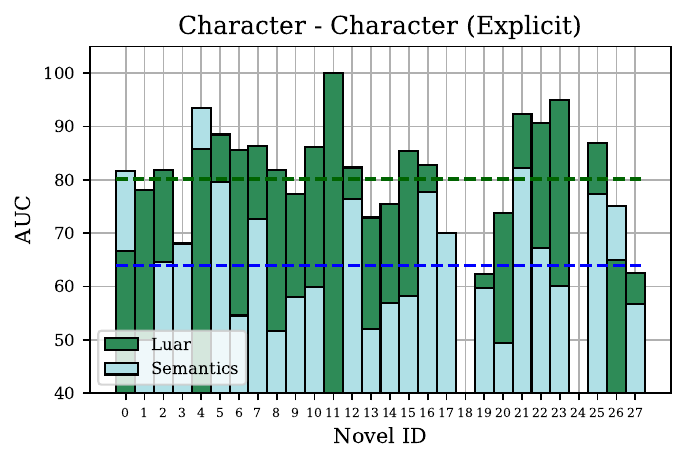}
    }
    \hfill
    \subfigure{\includegraphics[width=0.49\linewidth]{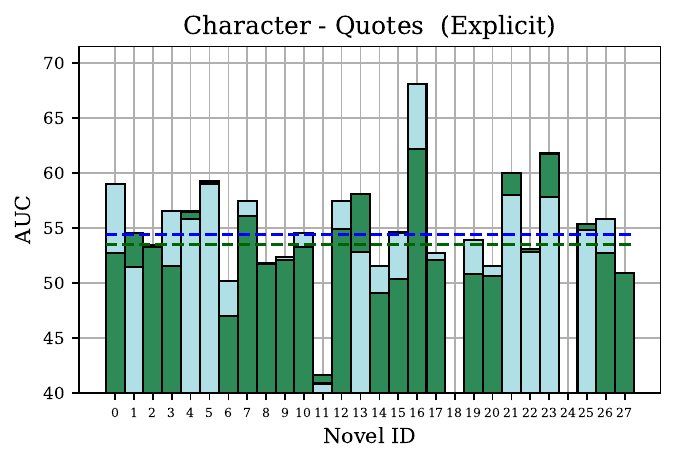}
    }
\caption{AUC per novel for the \textit{Explicit} experiment.}
\label{fig:explicit_novel}
\end{figure*}

We display the performance by novel for the Chapterwise experiment in Figure~\ref{fig:chapterwise_novel} and for the Explicit experiment in Figure~\ref{fig:explicit_novel}.
Note that for the Explicit experiment, novels 18 and 24 (\textit{The Gambler} by Fyodor Dostoevsky (1887) and \textit{The Sport of the Gods} by Paul Laurence Dunbar (1902) respectively) were not considered because we could not build queries due to the lack of explicit quotes. 
For the chapterwise experiment in the CQ evaluation setup, we see that LUAR's performances is higher than SBERT in 4 novels, indicating complementarity between these models.
The picture is even more evident in the explicit experiment (CQ setup), where LUAR's outperformns SBERT in 8 novels.
Overall, some novels exhibit characters voices where style information have more impact than on other novels. 

\section{Performance per Character Role}
\label{sec:appendix_B}

Table~\ref{tab:per_role} displays results of the Chapterwise and Explicit experiments by character role.
For the CC evaluation setup, LUAR performs very well on major characters, but struggles with intermediate characters.
On the other hand, the semantic-only model performs better on intermediate characters.
These results suggest complementarity between the two models, and that major characters exhibit more stylistic variations among them than intermediate characters.
The latter result can be linked to the authorial process of creating memorable major characters, with more unique voices than intermediate characters.

For the CQ evaluation setup, it seems that all models are better at attributing quotes of intermediate characters, and we see a quite large gap between the two roles. 

\section{Computing information}

We encode quotes with models on a 32-core Intel Xeon Gold 6244 CPU @ 3.60GHz CPU with 128GB RAM equipped with 3 RTX A5000 GPUs with 24GB RAM each.
For each model tested, one GPU was enough to encode all quotes in the 28 novels.
In total, running the full experiments took around 5 minutes for the Semantics and STEL models, 10 minutes for the Emotions model, and 1 hour for LUAR. 

\end{document}